\documentclass[journal]{IEEEtran}
\IEEEoverridecommandlockouts

\usepackage[hidelinks]{hyperref}
\usepackage[cmex10]{amsmath}
\usepackage{amssymb,amsfonts}
\usepackage{dblfloatfix}
\usepackage{amsthm}
\newtheorem{definition}{Definition}

\usepackage{graphicx}

\usepackage{booktabs}
\usepackage[numbers,compress]{natbib}
\usepackage{orcidlink}
\usepackage{subcaption} 
\usepackage{multirow}
\usepackage{xcolor}

\newcommand{\ourmodel}{CAD}
\newcommand{\ourcd}{TieCD}
\newcommand{\ourreg}{Light2Reg}

\usepackage{soul}
\usepackage{xcolor} 

\sethlcolor{yellow} 

\begin{document}

\title{Compress-Align-Detect: onboard change detection from unregistered images}

\author{Gabriele Inzerillo,
        Diego~Valsesia, \IEEEmembership{Member, IEEE}, Aniello Fiengo,
        and~Enrico~Magli, \IEEEmembership{Fellow, IEEE}%
\thanks{G. Inzerillo, D. Valsesia, E. Magli are with Politecnico di Torino -- Department of Electronics and Telecommunications, Italy. email: \{name.surname\}@polito.it. A. Fiengo is with the European Space Agency. The activity was selected via the Open Space Innovation Platform (https://ideas.esa.int) as a Co-Sponsored Research Agreement/Study/Early Technology Development and carried out under the Discovery programme of, and funded by, the European Space Agency (contract number: 4000143591). The view expressed in this publication can in no way be taken to reflect the official opinion of the European Space Agency.}
}

\maketitle
\begin{abstract}
 Change detection from satellite images typically incurs a delay ranging from several hours up to days because of latency in downlinking the acquired images and generating orthorectified image products at the ground stations; this may preclude real‐ or near real‐time applications. To overcome this limitation, we propose shifting the entire change detection workflow onboard satellites. This requires to simultaneously solve challenges in data storage, image registration and change detection with a strict complexity constraint. In this paper, we present a novel and efficient framework for onboard change detection that addresses the aforementioned challenges in an end-to-end fashion with a deep neural network composed of three interlinked submodules: (1) image compression, tailored to minimize onboard data storage resources; (2) lightweight co‐registration of non-orthorectified multi‐temporal image pairs; and (3) a novel temporally-invariant and computationally efficient change detection model. This is the first approach in the literature combining all these tasks in a single end-to-end framework with the constraints dictated by onboard processing. Experimental results compare each submodule with the current state‐of‐the‐art, and evaluate the performance of the overall integrated system in realistic setting on low-power hardware. Compelling change detection results are obtained in terms of F1 score as a function of compression rate, sustaining a throughput of 0.7 Mpixel/s on a 15W accelerator.

\end{abstract}


\section{Introduction}\label{sec:intro}

Change Detection (CD) has been long investigated in the field of remote sensing because of its fundamental importance in addressing problems such as disaster management, urban planning, land cover usage and much more. 

Nowadays, satellites equipped with optical, multispectral or hyperspectral sensors continuously scan the Earth's surface, converting reflected sunlight into raw digital counts. Once received at the ground station, the data undergo a processing pipeline including radiometric calibration and geometric corrections to compensate for sensor and satellite attitude and orthorectification to remove any distortion due to perspective and Earth's curvature. The resulting products may be used for CD.

While this ground-based workflow is able to produce very accurate results, it also incurs significant latency (from a few hours up to days) and is constrained by the need of high bandwidth links for downloading all images. At the same time, there is a growing interest in the remote sensing community towards moving inference tasks directly onboard satellites. New generation satellites begin to support lightweight deep learning (DL) models \cite{on-board_demonstrator, meoni2024unlocking, marin2021phi} for tasks such as segmentation and classification \cite{ziaja2021benchmarking, yao2019board, inzerillo2024efficient} to be performed directly in-orbit.
Onboard processing drastically reduces the data volume to be transmitted, alleviating bandwidth constraints and minimizing end‐to‐end latency, ultimately enabling truly real‐time monitoring. Such capabilities are especially important in the CD domain for time‐sensitive scenarios such as the rapid identification of natural disasters guiding to immediate damage assessment and recovery efforts.

However, onboard CD is still largely unexplored as it requires profound integration between multiple tasks and calls for novel frameworks in the way images are captured, stored and processed onboard. In particular, three system-level tasks need to be addressed: image compression and storage, image registration and detecting change. First, the satellite needs to store and retrieve images acquired at multiple revisits in order to detect change between pairs of them. If the area of interest is large, e.g., for worldwide coverage, the storage requirement can exceed the limited onboard resources, so compressing the images as efficiently as possible without compromising the ability to detect change accurately is crucial. Then, images available onboard are non-orthorectified products which typically present geometric disparities between revisits. While implementing the full orthorectification pipeline onboard is not feasible also due to the unavailability of auxiliary information such as digital elevation models, we need to ensure that image pairs to be used for CD are registered to each other. Finally, an efficient CD model is needed to output a change map.

Indeed, existing efforts towards onboard CD \cite{ruuvzivcka2022ravaen,xing2023lightcdnet,li2023lightweight,codegoni2023tinycd} only focus on lightweight neural networks operating on registered and orthorectified images, while neglecting registration and onboard storage. A few recent works \cite{zhao2023end, jing2025changerd} have proposed CD models for unregistered images. Although these approaches effectively tackle the challenge of detecting changes between misaligned image pairs, they are not designed with onboard deployment in mind. Specifically, they neglect considerations of computational efficiency, featuring significantly higher model complexity than ours, as well as the need to minimize onboard storage requirements.

In addition, we also observe that most deep learning CD models overlook the importance of temporal ordering of input image pairs, with the exception of only a few studies \cite{zheng2021change, zheng2022changemask, zheng2024segment, fang2023changer}. CD models are usually trained using image pairs presented in a fixed chronological sequence which biases them toward detecting changes in one specific temporal direction (for example, mainly building construction or mainly building demolition). Consequently, simply swapping the order of the two images at inference time can lead to dramatically degraded predictions. 
For real usage, an onboard CD model must therefore be agnostic to input ordering and capable of identifying any type of change, without being tied to a predetermined temporal direction.

In this paper, we thus present CAD (\textbf{C}ompress-\textbf{A}lign-\textbf{D}etect), an end-to-end framework for onboard CD that within a single modular neural network handles image compression, registration, and change detection. A single neural network model handling the whole CD pipeline allows to optimize it end-to-end, i.e., to maximize the change detection accuracy under a compression bitrate constraint, rather than optimizing the visual quality of the compressed images. This is clearly advantageous over a system composed of independent modules. Our contributions can be summarized as follows:
\begin{itemize}
    \item we design a lightweight end-to-end framework for onboard CD with three dedicated modules for image compression, co-registration, and change detection, each optimized for onboard deployment;
    \item we introduce \ourreg\ (\textbf{Light}weight \textbf{Reg}ressor for \textbf{Reg}istration), an efficient and lightweight image co-registration model;
    \item we proposed a novel lightweight CD model: \ourcd\ (\textbf{T}emporally-\textbf{I}nvariant and \textbf{E}fficient \textbf{C}hange \textbf{D}etector). \ourcd\ is architecturally invariant to temporal ordering, ensuring robust detection of changes, without biasing on temporal order of input images. In its larger configuration, it matches state-of-the-art performance with a limited computational cost;
    \item we provide a comprehensive evaluation of each submodule against corresponding baselines and then show, for the first time, the performance of the end-to-end onboard framework in terms of F1 detection score at a given compression bitrate;
    \item we provide experiments on low-power hardware demonstrating promising inference speed and memory requirements.
\end{itemize}

A preliminary version of this work appeared in \cite{inzerillo2025igarss}. Compared to the earlier version, we significantly innovate the design of the change detection, registration ad compression modules and significantly expand the experimental assessment.

\section{Background}

This work extends beyond CD to integrate several different image processing tasks such as image compression and co‐registration, that are necessary for enabling onboard CD. In the following section, we briefly review the related literature for each component, highlighting both state-of-the-art methods and, where possible, also lightweight and efficient approaches suitable for on-edge deployment.

\subsection{Image compression}

Image coding standards such as CCSDS 122.0‑B‑1 \cite{ccsds} and the more recent and widely adopted CCSDS 123.0‑B‑2 \cite{ccsds123} (for multispectral and hyperspectral imagery) are the de-facto choice in spaceborne missions. The CCSDS 122.0-B-1 standard employs a wavelet transform followed by bit-plane encoding and is specifically designed for onboard use with low hardware complexity. CCSDS 123.0‑B‑2 extends its predecessor by enabling near‑lossless compression with a predictive coding scheme. However, despite its broad adoption, the latter exhibits suboptimal rate–distortion performance in ultra‑low bitrate regime, which is a desirable operating point when images are used not for visual inspection but only to detect change. 

DL-based compression methods try to overcome this limitation by learning adaptive non-linear representations and dynamically allocating bits to optimize distortion or perceptual fidelity. 
The earliest methods \cite{balle2016end, balle2018variational} adopted autoencoder architectures using convolutional neural networks (CNNs) to obtain a compact latent representation, which is then quantized and entropy-coded at a given rate. Cheng et al. \cite{cheng2020learned} replaced classical entropy models with discretized gaussian mixture models and incorporated attention modules into the network, providing improved rate-distortion performance comparable to newer traditional compression methods, such as VVC \cite{bross2021overview}. He et al. \cite{he2022elic} observed that some channels contain much more information than others and proposed an unevenly grouped channel-wise context model for entropy coding. Although DL-based compressors have been extensively studied, there are few works specific for onboard compression \cite{pilikos2024raw}, since in such settings the limited computational resources make the design even more difficult. Valsesia et al. \cite{valsesia2024onboard} introduced a DL predictive compression architecture that processes hyperspectral data line-by-line, bounding both memory usage and computational complexity, and outperforming the CCSDS 123.0-B-2 standard.

\subsection{Image co-registration}

Image co-registration is the process of spatially aligning two images depicting the same scene, or subject, captured from different angles, at different times or even with different modalities. Classical techniques like SIFT \cite{sift} typically detect and describe salient keypoints, which are then used to find correspondences between two views and estimate a geometric transform to align the images. While effective in many scenarios, these methods can struggle under severe appearance changes, repetitive textures, or low‐contrast regions, and they often require careful parameter tuning or manual point selection. End‐to‐end DL networks like SuperGlue \cite{sarlin2020superglue} introduced graph neural networks over keypoint descriptors to establish correspondences that are significantly more accurate and robust to viewpoint and illumination variations than classical matching algorithms. Building on the foundation of SuperGlue, newer models such as LightGlue \cite{lindenberger2023lightglue} offer comparable matching accuracy while reducing computational requirements and inference time; however, they may still be too complex for effective onboard deployment. Ye et al. \cite{ye2022multiscale} proposed a multiscale framework for remote sensing multimodal registration trained in an unsupervised fashion, achieving accurate performance. Zhou et al. \cite{registration_change_detection} proposed a unified network to perform CD and image co-registration, inspired by the finding that both image registration and CD focus on extracting discriminative features. In this work we also perform them jointly, although our approach is quite different from \cite{registration_change_detection}, as explained in Sec. \ref{sec:method}.

\subsection{Change Detection}

Traditional CD methods go back much earlier than the development of DL-based techniques; early methods relied on direct pixel‐level comparisons, such as simple image differences or ratios, to highlight temporal variations in reflectance \cite{coppin1996digital}. Handcrafted feature approaches used change vector analysis to measure the magnitude and direction of spectral shifts between image pairs \cite{johnson1998change}, and principal component analysis for unsupervised CD \cite{celik2009unsupervised}. Object‐based techniques segmented imagery into homogeneous regions before comparing them to label the changed pixels \cite{walter2004object, desclee2006forest}. Traditional machine learning techniques such as random forests or support vector machines improved robustness but still suffered from shallow feature representation incapable of capturing complex patterns \cite{huang2008use, im2005change}. A huge paradigm shift occurred with DL and vision models: Daudt et al. \cite{daudt2018fully} made a big step forward by introducing the use of Siamese CNNs, originally designed for segmentation, to simultaneously process bitemporal images through parallel encoder branches. Other works have further improved by using new advanced DL layers in the models, such as attention mechanisms and transformers \cite{chen2020spatial, bandara2022transformer, fang2023changer}. The current state-of-the-art is undoubtedly represented by the foundation models \cite{wang2024mtp, li2024new}, which are first pretrained on massive remote sensing datasets and then finetuned to perform CD, among other tasks. While foundation models offer excellent performance, they are too complex for resource-constrained scenarios such as onboard processing, which led to the development of lightweight and efficient solutions. LightCDNet \cite{li2023lightweight} and TinyCDNet \cite{codegoni2023tinycd}, just to name a few, are two notable examples of lightweight models but with performance comparable to the state-of-the-art.

In Sec. \ref{sec:intro} we mentioned how important the property of temporal invariance is, i.e., for a CD model to always produce the same output regardless of the order in which the two input images are given to the model; only few works \cite{zheng2021change, zheng2022changemask, zheng2024segment, fang2023changer} explicitly target this property (referred to as \textit{temporal symmetry} in these works), promoting it in a soft manner by randomly swapping the input pair during training and using a loss that penalizes differences between the output for the original and the swapped order. Our approach is different: in Sec. \ref{sec:method} we introduce our novel CD architecture that is mathematically temporally-invariant: its internal operations guarantee identical outputs for both possible input orderings, without relying on augmented training or specialized loss functions.

\section{Proposed Method}\label{sec:method}

\begin{figure*}[t]
\centering
\includegraphics[width=1\linewidth]{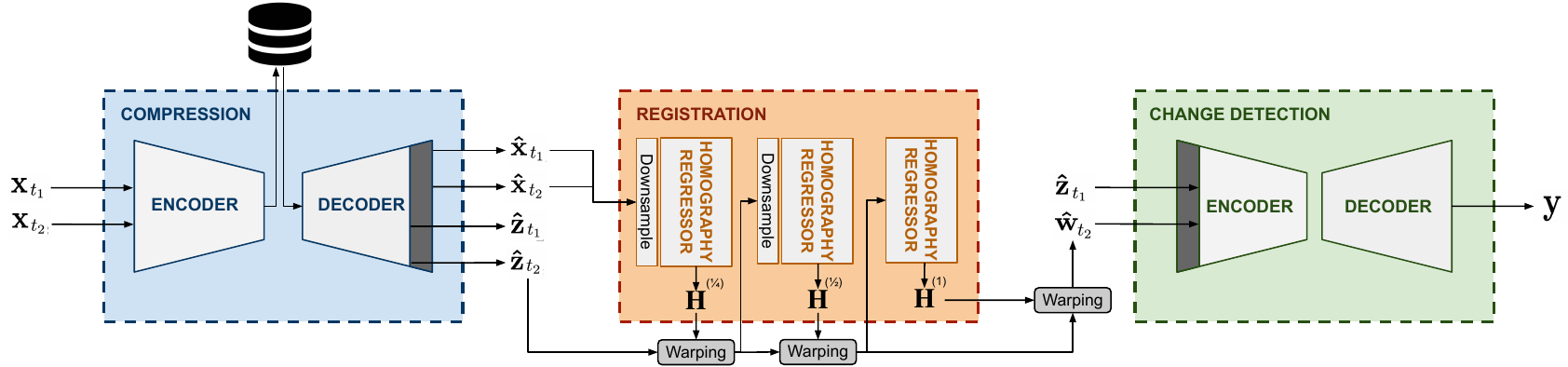}
\caption{The proposed \ourmodel\ framework takes as input a non-co-registered image pair, \( \mathbf{x}_{t_1} \) and \( \mathbf{x}_{t_2} \), acquired at different times. In a realistic onboard scenario, the earlier image \( \mathbf{x}_{t_1} \) is stored in compressed form and subsequently reconstructed via the decoder of the compression module; however, for clarity, we depict both images as parallel inputs. The reconstructed images, \( \mathbf{\hat{x}}_{t_1} \) and \( \mathbf{\hat{x}}_{t_2} \), are then passed to the co-registration module, which sequentially regresses three increasingly accurate homography matrices \(\mathbf{H}^{(1/4)}, \mathbf{H}^{(1/2)}, \mathbf{H}^{(1)}\). The final homography \( \mathbf{H}^{(1)} \) is used to warp the feature map \( \mathbf{\hat{z}}_{t_2} \) extracted from the decoder of the compression module onto the grid of \( \mathbf{\hat{z}}_{t_1} \). The aligned features are then passed to the CD module, \ourcd, which outputs the final change map \( \mathbf{\hat{y}} \). 
} 
\label{fig:overall_architecture}
\end{figure*}

The proposed end-to-end \ourmodel\ framework is illustrated in Fig.\ref{fig:overall_architecture}, which clearly outlines the three core submodules employed in this work. We begin by presenting a high-level overview of the entire framework, detailing the interaction between the individual components. We then dive into each model, highlighting the specific contributions and design choices that make them both effective and suitable for onboard deployment. Lastly, we explain the training procedure that we adopt to train the whole framework.

\subsection{Overall Framework}\label{subsec:framework}

The proposed \ourmodel\ framework consists of three modules: image compression, registration, and CD. Each of these components plays a crucial role in enabling efficient onboard change detection.

Given an input image pair \( \mathbf{x_{t_1}}, \mathbf{x_{t_2}} \in \mathbb{R}^{C \times H \times W} \), the framework can be summarized as follows:

\begin{align*}
    \mathbf{\hat{x}_{t_1}}, \mathbf{\hat{z}_{t_1}} &= \mathrm{Comp}(\mathbf{x_{t_1}}) \\
    \mathbf{\hat{x}_{t_2}}, \mathbf{\hat{z}_{t_2}} &= \mathrm{Comp}(\mathbf{x_{t_2}}) \\
    \mathbf{H^{(1)}} &= \mathrm{Reg}(\mathbf{\hat{x}_{t_1}}, \mathbf{\hat{x}_{t_2}}) \\
    \mathbf{\hat{w}_{t_2}} &= \mathrm{Warp}(\mathbf{\hat{x}_{t_2}}; \mathbf{H^{(1)}}) \\
    \mathbf{\hat{y}} &= \mathrm{CD}(\mathbf{\hat{z}_{t_1}}, \mathbf{\hat{w}_{t_2}})
\end{align*}

where $\mathrm{Comp}$ identifies the compression module, $\mathrm{Reg}$ the registration module and $\mathrm{CD}$ the change detection module. For each input image, the compression module produces two outputs: \( \mathbf{\hat{z}} \in \mathbb{R}^{F \times H/2 \times W/2} \) representing intermediate feature maps extracted from the penultimate layer of the decoder neural network, and $\hat{x}$ representing the reconstructed image. In the above notation, \( \mathbf{\hat{w}_{t_2}} \) is the spatially-warped version of \( \mathbf{\hat{z}_{t_2}} \) using the homography matrix \( \mathbf{H^{(1)}} \) calculated by the registration module. Finally, $\mathbf{\hat{y}}$ are the predicted change maps.

Onboard CD requires the satellite to store images until either the next revisit of a location of interest or for as long change might want to be detected. Since onboard storage is limited, we seek to optimize its usage for the CD task, i.e., spending the least bitrate needed to ensure good CD performance. This motivates the use of a neural image compression module, which, when end-to-end trained with the rest of the framework, allows to optimize the encoded latent representation for CD accuracy rather than simple visual quality. Each input image \( \mathbf{x} \) is compressed by an encoder generating a compact latent representation, which is then quantized, entropy coded and stored in a long-term onboard storage system. When needed for CD, this representation is fetched from storage matching metadata about the geographic location and decoded to reconstruct both the full-resolution image \( \mathbf{\hat{x}} \). Notice that the feature map output \( \mathbf{\hat{z}} \) from the penultimate layer  is also returned to enable a faster pipeline for registration and change detection that does not require decoding all the way back to pixel space.

After decompression, image co-registration is handled by the proposed registration module, called \ourreg, which receives the reconstructed image pair \( \mathbf{\hat{x}_{t_1}} \) and \( \mathbf{\hat{x}_{t_2}} \) and regresses a $3 \times 3$ homography matrix \( \mathbf{H^{(1)}} \). This matrix represents a projective spatial transformation (including translations, rotations, scaling, skewing, and perspective distortion) which is then used to warp the spatial grid of feature map \( \mathbf{\hat{z}_{t_2}} \) over the grid of \( \mathbf{\hat{z}_{t_1}} \), producing \( \mathbf{\hat{w}_{t_2}} \). Unlike classical registration pipelines, which are based on complex keypoint detection and iterative matching algorithms, this approximate alignment method is very computationally efficient and can be optimized end-to-end, making it suitable for the subsequent CD task. As shown in Fig. \ref{fig:overall_architecture}, full-resolution images are only reconstructed to compute the homography matrix, but warping and subsequent change detection are both performed directly in feature space, which is essential to make the framework faster and more efficient.

\subsection{Image compression module}

The modular design of our \ourmodel\ framework allows to use any state-of-the-art compression neural network, providing it meets complexity constraints and can be optimized end-to-end with a Lagrangian cost including rate and a task-specific loss (e.g., distortion, or, in our case, CD cross-entropy). 

In our experiments, we used as compression module a custom version of the well-known Scale Hyperprior architecture \cite{balle2018variational}. The model has a consolidated structure for compression tasks consisting of two encoders: a main encoder generates feature representations from input images, while a hyperprior entropy model learns the data distribution of input images. After the quantization step and the entropy coding, a decoder reconstructs the input. Although it does not achieve the rate–distortion performance of current state-of-the-art methods, its reconstruction quality remains more than adequate at a relatively small complexity, representing a balanced trade-off between compression quality and efficiency. In Sec. \ref{sec:results} we prove this claim by comparing it with newer state-of-the-art compression architectures such as ELIC \cite{he2022elic}.

\subsection{Image co-registration module}

\begin{figure}[t]
\centering
\includegraphics[width=1\linewidth]{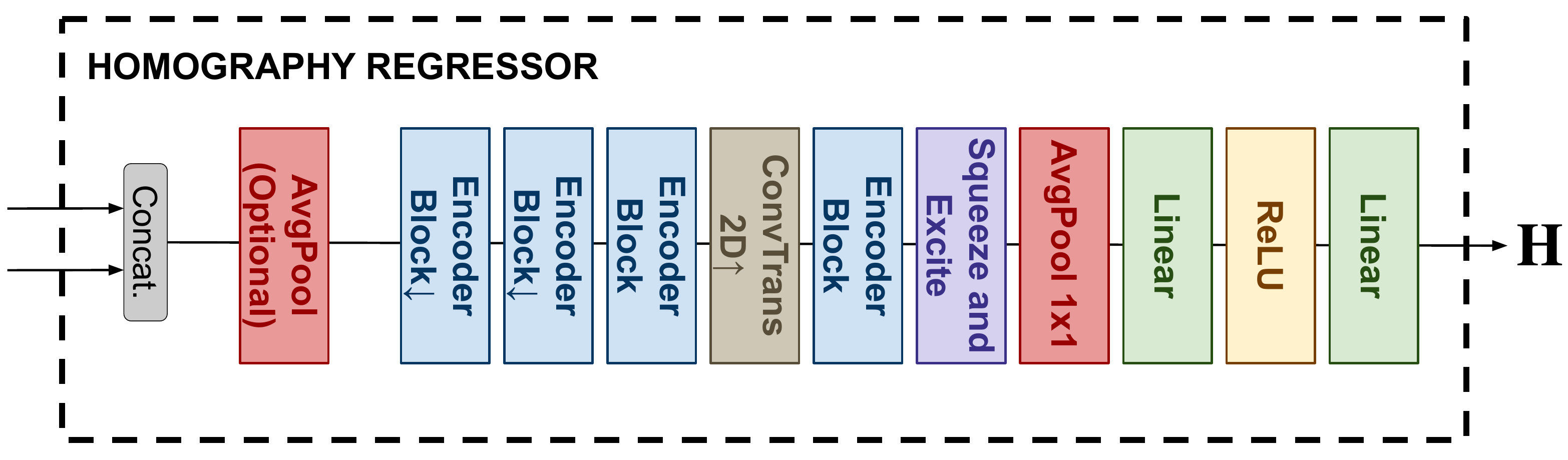}
\caption{Design of the homography regressor in \ourreg. After concatenating the two input images an optional average pooling can be performed. The encoder block, the main convolutional component of the regressor, is illustrated in Fig. \ref{fig:light2reg_encoder}. } 
\label{fig:light2reg}
\end{figure}

\begin{figure}[t]
\centering
\includegraphics[width=0.80\linewidth]{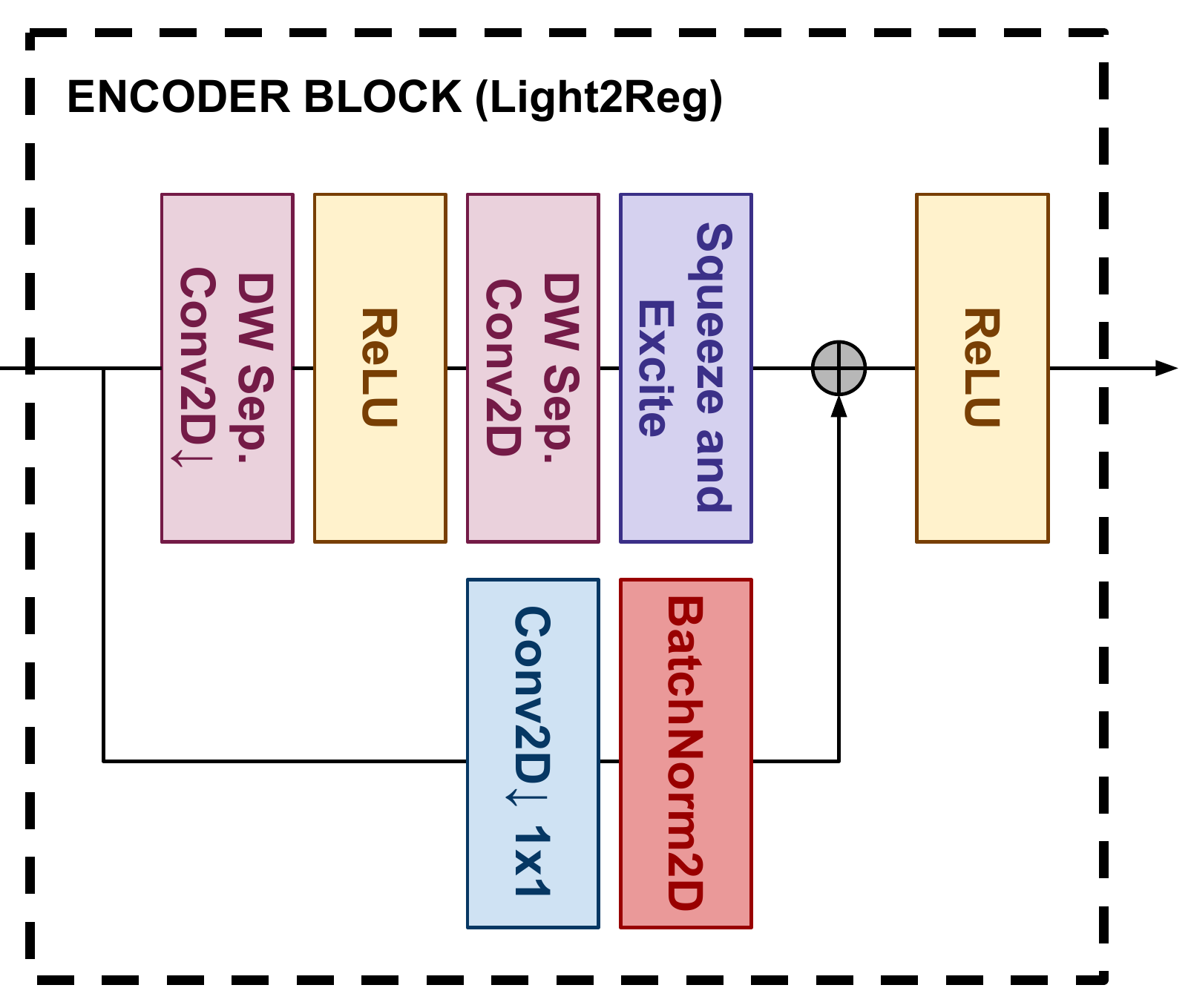}
\caption{Design of the efficient encoder block employed in \ourreg. Depthwise separable convolutions and just one \(1\times1\) 2D convolution make this component extremely light. } 
\label{fig:light2reg_encoder}
\end{figure}

DL–based co‐registration methods that achieve the highest accuracy typically rely on keypoint extraction and matching (e.g., SuperPoint \cite{detone2018superpoint} + SuperGlue \cite{sarlin2020superglue}), but these pipelines are computationally intensive, involve non‐differentiable matching steps, and cannot be trained end‑to‑end. To enable a lighter, fully differentiable approach, we approximate the geometric transform between image pairs with a direct homography regression. Our lightweight co-registration model, named \ourreg, follows a coarse-to-fine strategy, inspired by the structure of \cite{ye2022multiscale}, using a multiresolution cascade of three lightweight homography regressors (depicted in Fig.~\ref{fig:light2reg}). These regressors operate at three different spatial resolutions: \(1/4\), \(1/2\), and full resolution.  

We begin by downsampling the reconstructed images by a factor of 4 and processing them with a small CNN to extract coarse features, from which we predict an approximate homography matrix at quarter resolution, i.e., \(\mathbf{H}^{(1/4)}\). The homography regressor, and its internal encoder block, are depicted in Fig. \ref{fig:light2reg} and Fig. \ref{fig:light2reg_encoder}. It first concatenates the input images along the channel dimension and then uses a sequence of convolutional layers. Notice that it uses efficient separable convolutions to keep complexity low and channel attention \cite{hu2018squeeze} (``Squeeze and Excite'' block) to introduce input adaptivity. 

The estimated \(\mathbf{H}^{(1/4)}\) transformation is applied to the feature maps in the \( \mathbf{\hat{z}}_{t_2} \in \mathbb{R}^{F \times H/2 \times W/2} \) tensor extracted from the penultimate layer of the compression decoder. The warped features are then downsampled by a factor of 2 and processed by a second homography regressor with the same architecture but a higher number of features to extract a finer homography matrix \(\mathbf{H}^{(1/2)}\). This is used to further warp the feature maps and a third pass uses the last regressor, again with a higher number of features, to correct any remaining registration errors down to the pixel level and generating the last homography matrix \(\mathbf{H}^{(1)}\). This matrix is used for the final warping of the feature maps of the second image onto the spatial of grid of \( \mathbf{\hat{z}}_{t_1} \in \mathbb{R}^{F \times H/2 \times W/2} \), producing feature maps \( \mathbf{\hat{w}}_{t_2} \in \mathbb{R}^{F \times H/2 \times W/2} \). Tensors \( \mathbf{\hat{w}}_{t_2} \) and \( \mathbf{\hat{z}}_{t_1} \) serve as the input to the change detection module. Since they already represent downscaled image features, the CD module can skip some layers, usually dedicated to shallow feature extraction, reducing complexity.

\begin{figure}
\centering
\includegraphics[width=0.90\columnwidth]{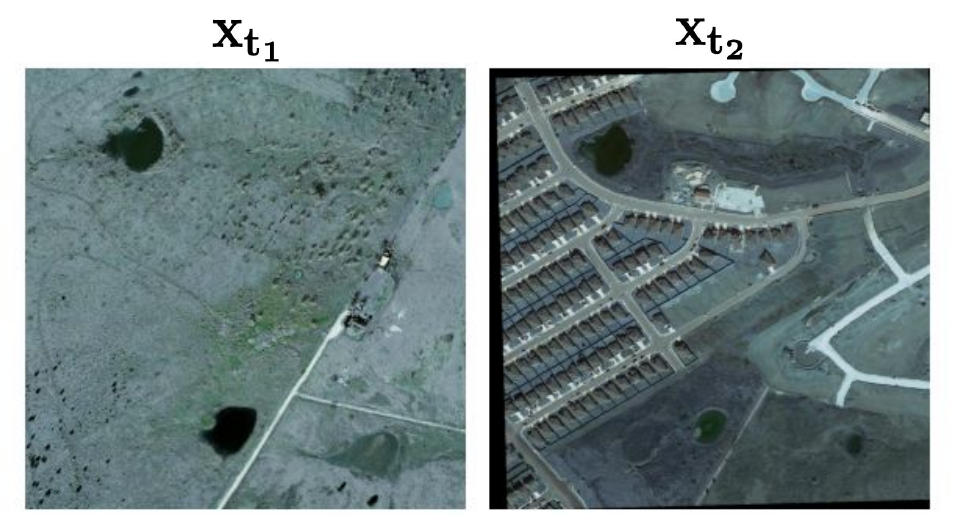}
\caption{Unregistered transformed image pair. In addition to the geometric distortions that misalign the images, the semantic content is significantly different: note how \(\mathbf{x}_{t_1}\) contains many more elements than \(\mathbf{x}_{t_2}\) and how little remains in common.}
\label{fig:registration_src_ref}
\end{figure}

One important remark is to be made about image registration. We assume that the geometric relationships between images acquired at multiple revisits can be approximated by a projective transformation. This is, in general, suboptimal and more careful registration models could be devised if additional side information was available such as elevation models, etc.. However, it is unrealistic to assume that such information is available or usable with low complexity onboard. Moreover, we remark that, to the best of our knowledge, there are currently no public large datasets that provide non-orthorectified, unregistered satellite images for change detection. Therefore, our experiments rely on simulations based on random projective transformations to the input image pairs. Although this synthetic approach can never fully reproduce real onboard data, applying strong distortions to images is a common technique in the remote sensing image co-registration literature \cite{ye2022multiscale, registration_change_detection,  zhu2023advances, ye2019fast}, ensuring that the reference and source data are strongly misaligned and making the co-registration task anything but trivial. Notice that since we are in the CD setting, the geometric transformations are applied to multitemporal images where the content variations make registration more complex. An example from our testing data is shown in Fig. \ref{fig:registration_src_ref} where significant change has occurred to the same area in addition to the geometric transformation. 

\subsection{Change detection module}
\label{subsec:cd_module}

\begin{figure*}[t]
\centering
\includegraphics[width=0.85\linewidth]{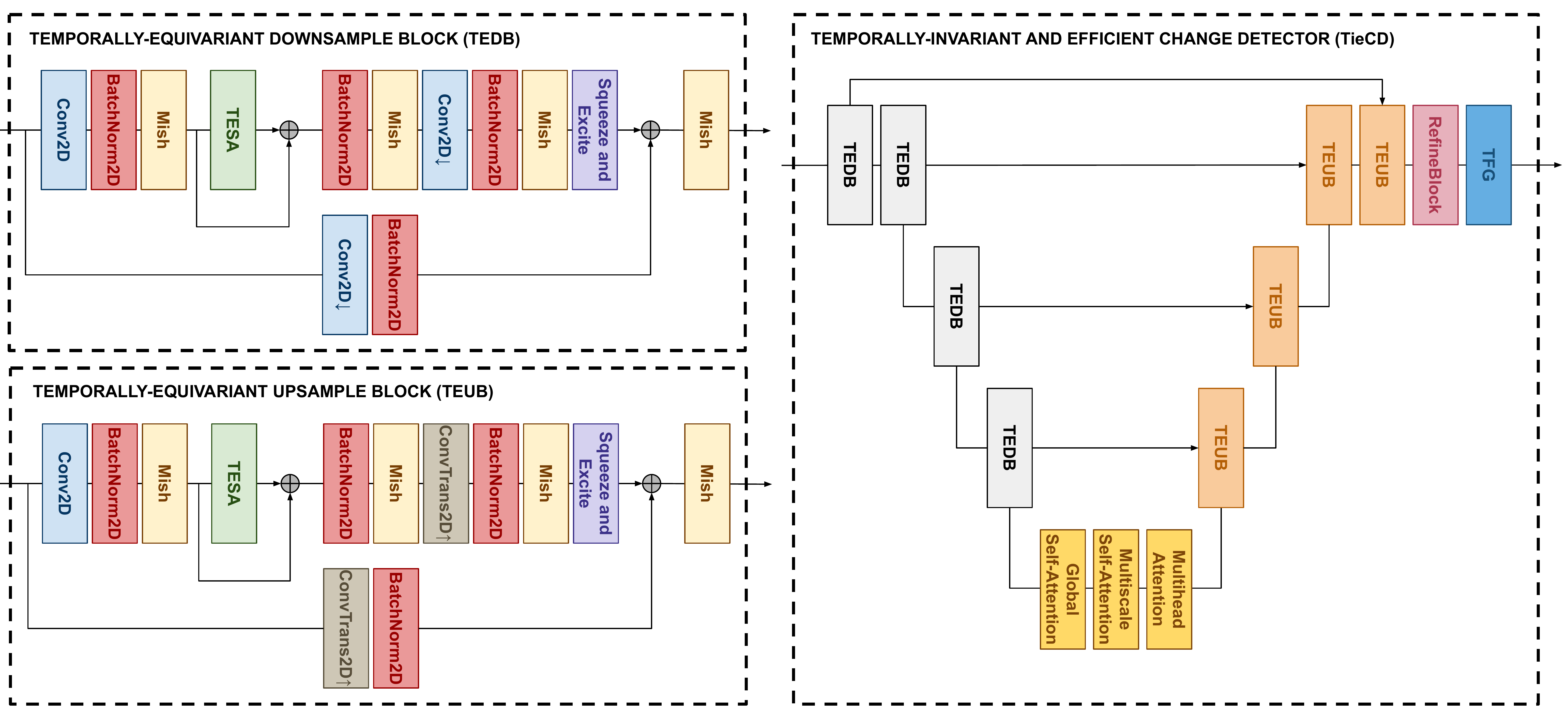}
\caption{\textit{Left}: designs of the Temporally-Equivariant Downsample Block (TEDB) and Temporally-Equivariant Upsample Block (TEUB), used as main building blocks of TieCD. \textit{Right}: design of Temporally-Invariant and Efficient Change Detector (\ourcd). By concatenating the input images on the batch dimension and performing a series of equivariant operations on the order of permutations we maintain the equivariance property. Invariance is eventually achieved using a global temporally-invariant operation: the Temporal Fusion Gate (TFG).} 
\label{fig:tiecd}
\end{figure*}

Beyond efficiency, crucial for onboard deployment, our CD model is built from the driving idea that the model itself should be invariant to temporal permutations of the input image pair, i.e., the change map should be the same irrespective of the ordering of the two images. This property is often overlooked in several state-of-the-art designs which leads to overfitting the temporal order. Especially when limited training data are available, a non-invariant model may overfit a particular ordering dominant in the training set (e.g., building construction rather than demolition) leading to poor performance on the reverse ordering. Our experimental results prove that several state-of-the-art models degrade to unusable levels when the ordering of the input images is reversed. Some works \cite{zheng2021change, zheng2022changemask, zheng2024segment, fang2023changer} acknowledge the importance of the temporal invariance property, and tackle it by training-time augmentation where the ordering of image pair is randomly swapped during training and by using losses that minimize the error given by both orderings \((\mathbf{x}_{t_1},\mathbf{x}_{t_2})\) and \((\mathbf{x}_{t_2},\mathbf{x}_{t_1})\). These solutions can improve a model's robustness to input order, and may be effective in some cases, but they only provide a soft invariance. Furthermore, they attempt to learn the desired invariance from the data themselves, rather than encoding it in the design, possibly leading to a suboptimal use of data, especially when their quantity is limited.

\begin{figure}
\centering
\includegraphics[width=0.9\columnwidth]{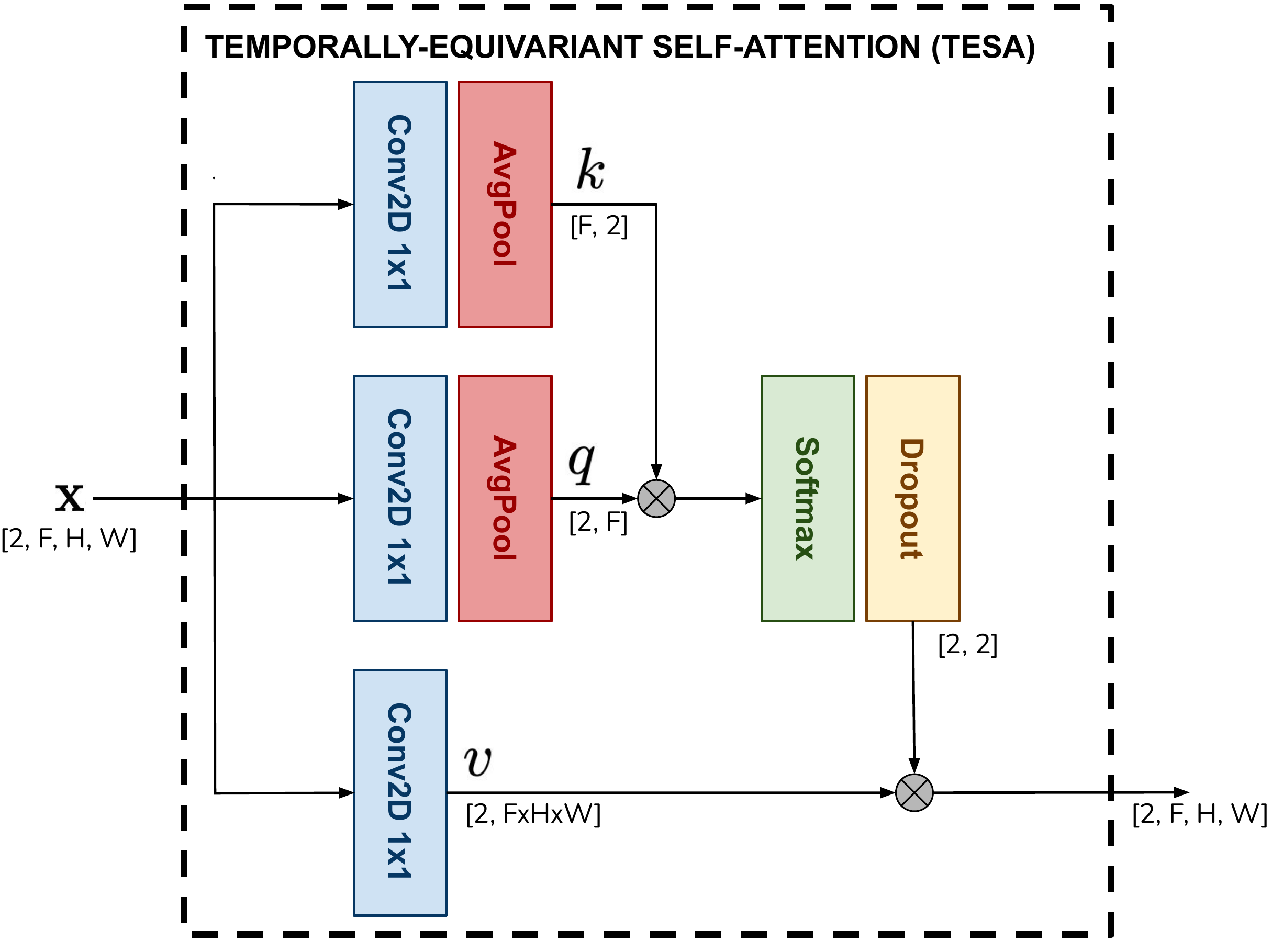}
\caption{Design of the temporally-equivariant self-attention (TESA) block. It computes pairwise attention scores within each temporal pair, swapping the input pair ($\mathbf{x}_{t_1}\leftrightarrow \mathbf{x}_{t_2}$) simply permutes the outputs in the same exact way, satisfying the formal definition of equivariance.}
\label{fig:tesa}
\end{figure}

Our proposed Temporally-Invariant and Efficient CD (\ourcd) module represents a lightweight architecture for onboard CD which also guarantees strict mathematical invariance by design, regardless of training data, training procedure or losses. 
We achieve temporal invariance through the usage of equivariant and invariant functions, whose definitions are given below. 

\begin{definition}[Invariance]
\textit{A function \( f : X \rightarrow Y \) is said to be invariant to the actions \( g \) of a group \( \mathcal{G} \) if}
\[
f(g \circ x) = f(x) \quad \text{for all } x \in X, \, g \in \mathcal{G}.
\]
\end{definition}

\begin{definition}[Equivariance]
\textit{A function \( f : X \rightarrow Y \) is said to be equivariant to the actions \( g \) of a group \( \mathcal{G} \) if}
\[
f(g \circ x) = g \circ f(x) \quad \text{for all } x \in X, \, g \in \mathcal{G}.
\]
\end{definition}

While the simplest way to design a model invariant to input ordering would be to use only invariant layers throughout the network, this approach has a major limitation: there are only a limited number of simple functions invariant to permutations, and they are often too simple to capture complex spatial features necessary for CD. To address this, we adopt a more flexible strategy: we design the architecture stacking layers that are equivariant to permutation, and at the very end we apply a global invariant operation, ensuring that the final output is strictly independent of the input order, while still allowing the network to model rich and expressive features. 

The proposed architecture is based on the classical U-Net structure \cite{ronneberger2015u} and it is  depicted in Fig. \ref{fig:tiecd}. Unlike traditional approaches, input images are not concatenated along the channel dimension, as any projection along this dimension would break equivariance. Instead, each network block either processes them in parallel with shared weights or mixes them with a permutation-equivariant attention operation. The main network blocks (Temporally-Equivariant Downsampling/Upsampling Blocks) are TEDB/TEUB illustrated in Fig. \ref{fig:tiecd}. All their operations are independently applied to each input image, except for the Temporal Equivariant Self-Attention (TESA) block that mixes the features of the two input images in a permutation-equivariant way to create joint representations. This block, depicted in Fig. \ref{fig:tesa}, applies a self-attention \cite{vaswani2017attention} to a length-2 sequence composed of the features of the two images; the output is again a length-2 sequence of features in which element has been modulated by the learned temporal relationships. Crucially, swapping the input pair simply permutes the outputs, preserving equivariance. More in detail, the feature maps of the two images undergo shared $1 \times 1$ convolutions and spatial averaging to form the key $\mathbf{k} \in \mathbb{R}^{2\times 2 F}$ and query matrices $\mathbf{k} \in \mathbb{R}^{2\times F}$. A $2\times2$ attention matrix is then derived to temporally mix the value matrix $\mathbf{v}$, also obtained by $1 \times 1$ convolution and vectorized to size $2 \times FHW$:
\begin{align}
    \mathbf{z} &= \mathrm{Softmax}\left( \frac{\mathbf{q} \cdot \mathbf{k}^T}{\sqrt{F}} \right) \mathbf{v}
\end{align}

In the bottleneck of our \ourcd\ U-Net, we stack three attention operations: a Global Self-Attention, a Multiscale Self-Attention and a final Multihead Attention. We remark that these attentions operate independently on each of the images in the spatial dimension, not temporally as the one previously introduced for the TESA block.

The Global Self-Attention block enhances features by computing long-range dependencies through spatial self-attention and applying a residual refinement. Key and query tensors of size $\frac{F}{8} \times H \times W$ are obtained by $1\times1$ convolution from the original $F$-dimensional feature maps to reduce computational cost and vectorized to $HW \times \frac{F}{8}$ to compute attention scores over all the pixels, while the value tensor is projected in the same way to an $F$-dimensional space. In formulas:

\begin{align*}
  \mathbf{A} &= \mathrm{Softmax}\left( \frac{\mathbf{q} \cdot \mathbf{k}^T}{\sqrt{F/8}} \right) \in \mathbb{R}^{HW \times HW} \\
  \mathbf{o} &= \mathbf{A}\mathbf{v} \in \mathbb{R}^{HW \times F}.
\end{align*}
After reshaping back to $F \times H \times W$ a separable convolution, BatchNorm and a Mish non-linearity \cite{misra2019mish} are applied to compute the output of the block.

The subsequent Multi-Scale Attention block enhances features by applying the same global self-attention previously explained at three different scales: the original, the one obtained after an undecimated averaging filter of size $2 \times 2$ and the one obtained after an undecimated averaging filter of size $4 \times 4$, stacked to form a tensor $\mathbf{o_{\text{cat}}}$ of size $3F \times H \times W$. This is then processed with 2D convolution, BatchNorm and Mish non-linearity.

Finally, the features $\mathbf{z} \in \mathbb{R}^{F \times H \times W}$ produced by the Multi-Scale Attention block undergo a final global attention operation in the Multihead Attention block.
\begin{align*}
\mathbf{z'} &= \mathrm{reshape}(\mathbf{z}) \in \mathbb{R}^{HW \times F} \\
\mathbf{h}_1 &= \mathbf{z'} + \mathrm{MHSA}(\mathrm{LN}_1(\mathbf{z'})) \\
\mathbf{h}_2 &= \mathbf{h}_1 + \mathrm{MLP}(\mathrm{LN}_2(\mathbf{h}_1)) \\
\mathbf{y} &= \mathrm{reshape}(\mathbf{h}_2) \in \mathbb{R}^{F \times H \times W}
\end{align*}
where \(\mathrm{MHSA}\) is the multi-head self-attention mechanism, \(\mathrm{MLP}\) is a two-layer linear network with GeLU activation, and \(\mathrm{LN}_i\) denotes layer normalization. 

The decoder portion of the temporally-equivariant part of the U-Net architecture is terminated by a Refinement Block, whose goal is to refine the feature maps. The refinement process is the following:
\begin{align*}
  \mathbf{r}_1 &= \mathrm{Mish}\left( \mathrm{BN}_1\left( \mathrm{Conv}_{5 \times 5}(\mathbf{z}) \right) \right) \\
  \mathbf{r}_2 &= \mathrm{Mish}\left( \mathrm{BN}_2\left( \mathrm{Conv}_{3 \times 3}(\mathbf{r}_1) \right) \right) \\
  \mathbf{y} &= \mathrm{Conv}_{1 \times 1}(\mathbf{r}_2)
\end{align*}

After the refinement block, each of the two input images has an $F$-dimensional feature representation for each pixel which will be the same regardless of the ordering of the input images. 

The last block of the network (Temporal Fusion Gate (TFG)) is concerned with fusing the features of the two images into a single feature space that is invariant to ordering. Let $\mathbf{z}_1, \mathbf{z}_2 \in \mathbb{R}^{F\times H\times W}$ denote the two feature maps produced for the input pair. TFG computes a fused feature map $f_{\mathrm{fused}}$ that is strictly invariant to the order of its inputs via the following sequence of operations:

\begin{align}
  \mathbf{a} &= \frac{1}{2}\bigl(\mathbf{z}_1 + \mathbf{z}_2\bigr) 
  \label{eq:avg}\\
  \mathbf{d} &= \bigl\lvert \mathbf{z}_1 - \mathbf{z}_2 \bigr\rvert
  \label{eq:diff}\\
  \mathbf{c} &= \frac{1}{2}\bigl(\mathbf{a} + \mathbf{d}\bigr)
  \label{eq:combine}\\
  \mathbf{m} &= \sigma\!\bigl(\mathrm{Conv}(\mathbf{c})\bigr)
  \label{eq:modulation}\\
  \mathbf{f_{\mathrm{fused}}} &= \mathbf{m} \odot \mathbf{a}
  \label{eq:fuse}
\end{align}
where Eq. \eqref{eq:avg} and Eq. \eqref{eq:diff} compute symmetric statistics via average and absolute difference, respectively, which are then merged by Eq. \eqref{eq:combine} into combined features $\mathbf{c}$. Finally, Eq. \eqref{eq:modulation} applies a convolutional layer followed by a sigmoid activation $\sigma$ to produce a modulation mask $\mathbf{m}\in[0,1]^{F\times H\times W}$. Finally, Eq. \eqref{eq:fuse} multiplies $\mathbf{m}$ with the average features $\mathbf{a}$ to yield the fused feature map. This allows to implement a spatially-adaptive fusion mechanism. Notice that it is easy to verify that $\mathbf{f}_{\mathrm{fused}}$ is invariant to the ordering of $\mathbf{z}_1$ and $\mathbf{z}_2$.

\subsection{Training Procedure}\label{subsec:training_procedure}

Training a complex and modular framework such as \ourmodel\ is far from trivial. Direct end-to-end optimization straight from random initialization may be suboptimal as it is hard for the sub-modules to learn the specifics of their functionality for the CD objective alone. Also, it is easy for gradients to vanish or explode over such a long pipeline. For these reasons, it is crucial to carefully design the training strategy, not just for each individual module, but for the framework as a whole.
We start by pretraining each module independently, optimizing it with its own respective loss function, which we now describe, in order to let it learn the desired functionality within the framework.

For the compression module we follow a well-established approach \cite{balle2016end, balle2018variational} and minimize the following rate-distortion loss:

\begin{equation}
\mathcal{L}_{\mathcal{C}} = \lambda R + D,
\label{eq:compression_loss}
\end{equation}
where
\begin{equation}
R = \mathbb{E}_{\mathbf{x}\sim p_{\mathrm{data}}}
\bigl[-\log p_{\mathbf{\hat{z}}}(\mathbf{\hat{z}})\bigr]
\label{eq:rate_eq}
\end{equation}
is the expected bitrate of the quantized latent code \(\mathbf{\hat{z}}\), and
\begin{equation}
D = \mathbb{E}_{x\sim p_{\mathrm{data}}}
\bigl\lVert \mathbf{x} - \widehat{\mathbf{x}} \bigr\rVert_{2}^{2}
\end{equation}
is the mean squared reconstruction error between the original image \(\mathbf{x}\) and its reconstruction \(\hat{\mathbf{x}}\). The hyperparameter \(\lambda\) balances rate and distortion, leading to different operating points.

Concerning the registration module, we recall that \ourreg\ generates three progressively more and more precise homographic matrices. The registration loss aggregates the misalignment errors at each of the three scales \(\in \lbrace 1/4,1/2,1 \rbrace\) by comparing the reference image patches \(\mathbf{x}_{t_1}^{(s)}\) to their warped counterparts \(\mathbf{w}_{t_2}^{(s)}\). Suppose that \(\mathbf{H}^{(s)}\) denotes the \(3\times3\) homography matrix predicted at resolution \(s\in\{1/4,1/2,1\}\), then
\begin{equation}
w_{t_2}^{(s)} = \mathrm{Warp}\bigl(x_{t_2}^{(s/2)};\,H^{(s)}\bigr)
\end{equation}
The total registration loss is then defined as
\begin{align}
\mathcal{L}_{\mathcal{R}} =\; &\alpha_{1}\,\bigl\lVert \mathbf{x}_{t_1}^{(1)} - \mathbf{w}_{t_2}^{(1)}\bigr\rVert_{2}^{2} \\
+\; &\alpha_{2}\,\bigl\lVert \mathbf{x}_{t_1}^{(1/2)} - \mathbf{w}_{t_2}^{(1/2)}\bigr\rVert_{2}^{2} \\
+\; &\alpha_{3}\,\bigl\lVert \mathbf{x}_{t_1}^{(1/4)} - \mathbf{w}_{t_2}^{(1/4)}\bigr\rVert_{2}^{2}\,
\label{eq:reg_loss}
\end{align}
Here, each weight \(\alpha_s\) balances the contribution of its corresponding scale. By penalizing squared differences at progressively finer resolutions, the network first corrects large misalignments on coarse grids and then refines them to subpixel precision at full resolution. To pretrain the registration network independently we minimize the loss (\ref{eq:reg_loss}) during the pretraining.

Lastly, to pretrain the CD module we optimize a standard cross-entropy loss:
\begin{equation}
\mathcal{L}_{\text{CD}} = - \sum_{i} \left[ y_i \log(\hat{y}_i) + (1 - y_i) \log(1 - \hat{y}_i) \right]
\label{eq:cd_loss}
\end{equation}
where \( y_i \in \{0, 1\} \) is the ground truth label indicating the presence or absence of change on pixel \( i \), and \( \hat{y}_i \in [0, 1] \) is the predicted probability. Notice that pretraining the CD module requires adding two layers at the beginning to let it operate in the full-resolution pixel space rather than in the downsampled feature space that we described in Sec. \ref{subsec:cd_module}. These layers are then removed for the integration in the pipeline.

After individually pretraining each module, we proceed with a joint pretraining of the registration and CD modules. This joint training of these two modules serves two main purposes. First, it improves the performance on both tasks which mutually benefit from the joint optimization, as we show in Sec. \ref{sec:results}. Second, it defines an upper bound on CD accuracy, for the entire framework when no compression is applied. This provides a reference point for evaluating the whole framework, including the compression module, to quantify the potential performance degradation introduced by compression, which we recall is necessary, as it significantly reduces onboard storage utilization. For the joint registration and CD pretraining (referred as \textit{Reg-CD}), we employ a composite loss function that combines the registration loss from Eq. \eqref{eq:reg_loss} and the CD loss defined in Eq. \eqref{eq:cd_loss}. The total loss is formulated as:

\begin{equation}
\mathcal{L}_{\text{RCD}} = \alpha \mathcal{L}_{\text{CD}} + (1 - \alpha) \mathcal{L}_{\text{R}}
\label{eq:reg_cd_loss}
\end{equation}
where \( \alpha \in [0,1] \) is a hyperparameter that balances the contributions of the change detection loss \( \mathcal{L}_{\text{CD}} \) and the registration loss \( \mathcal{L}_{\text{R}} \), encouraging the network to improve both alignment and detection accuracy simultaneously.

Finally, the whole framework, comprising compression, registration, and CD, is finetuned end-to-end. For this purpose, we adopt a rate-penalized Lagrangian objective that jointly optimizes the three modules. Specifically, we combine the change detection loss \( \mathcal{L}_{\text{CD}} \), the registration loss \( \mathcal{L}_{\text{R}} \), and the rate of the compressed image representation defined in Eq. \eqref{eq:rate_eq}. The total finetuning loss is defined as:

\begin{equation}
\mathcal{L}_{\text{total}} = \left[ \alpha \mathcal{L}_{\text{CD}} + (1 - \alpha) \mathcal{L}_{\text{R}} \right]+ \lambda R
\label{eq:full_loss}
\end{equation}

Looking at our loss for the whole framework, we notice how we initially pretrain the compression module using a conventional rate-distortion loss (as described in Eq.~\eqref{eq:compression_loss}), allowing us to learn an effective representation to encode the input images. However, unlike standard image compression pipelines that aim to reconstruct perceptually faithful images, our objective changes in the full end-to-end setting. Once the registration and change detection modules are also trained, we finetune the entire architecture jointly using the total loss. In this stage, we no longer optimize for reconstruction quality, instead, we treat both CD and registration performances as distortion measure. This shift reflects a \textit{coding-for-machines} \cite{scalable} approach, where the compression model is optimized to reconstruct the most relevant features to promote accurate CD, rather than generating visually faithful reconstructions of the input images.

\section{Experimental results}\label{sec:results}

In this section we present the experimental results of the entire framework, including the three modules, as well as an extensive analysis, concerning both performance and complexity, of each module. FLOPs and number of parameters reported in this section are computed using the \textit{calflops} \cite{calflops} package with input images of size $3 \times 512 \times 512$.

\subsection{Datasets}

To train and evaluate the effectiveness of the proposed method, and each submodule, we used two datasets for different purposes. We used CloudSEN12 \cite{aybar2022cloudsen12} to pretrain the registration network alone. LevirCD \cite{chen2020spatial} was used to perform compression pretraining, CD pretraining, joint \textit{Reg-CD} pretraining, end-to-end finetuning and evaluation of the entire framework. CloudSEN12 \cite{aybar2022cloudsen12} comprises over 49,000 multi-temporal image patches, spanning clear scenes, thick and thin clouds, cloud shadows, along with dense pixel-level annotations. Although its primary purpose is cloud segmentation, the diversity and scale of the dataset make it well suited for pretraining our registration module. Each image is labeled with a label that indicates the cloud coverage; to pretrain the registration network we used only the subset of images labeled as "low-cloudy", "almost-clear", "cloud-free" to avoid the registration of scenes containing only clouds. Furthermore, we just used RGB spectral bands, to maintain consistency with the LevirCD dataset. LevirCD \cite{chen2020spatial} is the most widely used benchmark for the CD task. It consists of 637 high-resolution (spatial resolution is 0.5 meters) $1024 \times 1024$ 8-bit RGB images. To pretrain the compression module, our CD model and jointly pretrain registration and CD networks (\textit{Reg-CD}) we followed the standard splits provided by the authors of the dataset, randomly cropping training images to $512 \times 512$ patches.

\subsection{Implementation details}\label{subsec:impl_details}

All models were implemented, and experiments were conducted, using PyTorch library for python. For training we used one NVIDIA A40-48GB GPU, while for evaluation under a low-power, resource-constrained setup, we used the NVIDIA Jetson Orin Nano, a commercial 8GB embedded system designed for AI inference. For the latter, we conducted our evaluation using the 15W power budget mode. For all the trainings we used the Adam optimizer \cite{Kingma2014AdamAM}. 

To simulate the lack of orthorectification and co-registration in images, we applied strong affine and perspective transformations using the \textit{torchvision} library. For training involving CD, the transformations are applied to one of the images in the image pair, while for pretraining of only the registration module we applied transformations to the single reference image to obtain the source image. It is important to remark that \ourmodel\ was trained randomly sampling different spatial transformations at each epoch in order to make it robust. For this reason, we want to test our proposed framework under a variety of random distortions in order to assess its robustness. Thus, we evaluate \ourmodel\ with 100 independent testing runs, each with a different set of distortions, and report error bars on the results related to this variability. Error bars in all figures report one standard deviation around the mean. In comparative experiments where we test different models, the same sets of distortions are applied to all the models, to guarantee a fair comparison.

We list here all the implementation details related to each evaluation and training setup:

\begin{itemize}
    \item \textit{Compression module}: compression pretraining and evaluation was performed on the LevirCD dataset, using the default train, evaluation and test splits. Learning rate was set equal to $10^{-4}$ for encoder-decoder model and $10^{-5}$ for the entropy bottleneck model. Models were trained for 400 epochs with a batch size of 8.
    Different $\lambda$ were used, leading to different models with different rate-distortion trade-offs. The chosen efficient compression module is a Scale Hyperprior \cite{balle2018variational}  with a custom number of features: 150 latent channels for the encoder-decoder part and 225 latent channels for entropy bottleneck. In the ablations, we assess other compression models (ELIC \cite{he2022elic}, Factorized Prior \cite{balle2018variational}) for which we used the implementations in the  \textit{compressai} library \cite{begaint2020compressai}.
    \item \textit{Registration module}: registration pretraining followed a classical unsupervised approach, we applied distortions to a source image from CloudSEN12 datasets and train \ourreg\ to regress the homography to warp source over reference image. Models were trained for 500 epochs with a batch size of 24 and a learning rate of $10^{-4}$.
    \item \textit{CD module}: we study two variants of \ourcd, with different levels of complexity, namely \ourcd-L and \ourcd-S. The former has a total complexity of 59 GFLops, while the latter is a highly compact variant using only half the number of features for a total of 15 GFlops. We pretrained them for 1500 epochs, with a batch size of 8, and $10^{-5}$ as starting learning rate which we decreased using a step scheduler (\(\gamma\)=0.3) after 1000 epochs. For the pretraining step, no distortions were applied to the image pairs. We used random rotation, flipping (both horizontal and vertical) and color jittering as training augmentations. To train and evaluate other baseline CD models, we used implementations from the OpenCD \cite{opencd} library. Notably, this implementation of many models achieves even better results than those reported by the original authors.   
    \item \textit{Joint Reg-CD}: to perform the joint pretraining of \ourreg\ and \ourcd\ we used as starting weights the ones obtained from the independent pretraining of each module. We minimized the loss in Eq. \eqref{eq:reg_cd_loss} by using \(\alpha = 0.3\). Since in this setting compression is not performed and thus feature maps from the compressor are not available, the two modules work with all inputs in the pixel space avoiding passing feature maps to \ourcd. For this experiment, batch size was set to 4, starting learning rate to $3\cdot 10^{-5}$, weight decay to $10^{-5}$, and the number of epochs to 1500. After 1000 epochs we multiplied the starting learning rate by 0.3.
    \item \textit{Entire Framework}: to finetune the whole \ourmodel\ framework we started from the pretrained weights obtained from the compression pretrain, for the compression model, and from the joint \textit{Reg-CD} weights for \ourreg\ and \ourcd. Since the entire framework performs compression, we conducted many experiments testing different $\lambda$ values and producing as many sets of weights for all the three submodules. Notice that without loss of generality, variable bitrate training techniques known in the literature could also be used. In the total loss (Eq. \ref{eq:full_loss}) we set \(\alpha=0.5\). For all experiments, batch size was set to 4, learning rate to $3\cdot 10^{-5}$, and number of epochs to 1000.
\end{itemize}

\subsection{Evaluation metrics}

Since we perform three different tasks to achieve onboard CD, in this work, we employ different quantitative metrics to evaluate not only the effectiveness of CD, but also compression and registration.

\paragraph{Peak Signal-to-Noise Ratio (PSNR)}  
to evaluate distortion in compression we report PSNR:
\begin{equation}
\mathrm{PSNR} \;=\;10\log_{10}\!\Bigl(\frac{L^{2}}{\mathrm{MSE}}\Bigr)
\end{equation}
where \(L\) is the maximum pixel value (e.g., 255 for 8‑bit images).

\paragraph{F1‐score}  
for CD, we focus on the positive (change) class and compute the micro‑averaged F\(_1\)‐score:
\begin{align}
\mathrm{Precision}= \frac{\mathrm{TP}}{\mathrm{TP} + \mathrm{FP}}\\
\mathrm{Recall} = \frac{\mathrm{TP}}{\mathrm{TP} + \mathrm{FN}}\\
F_1 = 2 \cdot \frac{\mathrm{Precision} \times \mathrm{Recall}}{\mathrm{Precision} + \mathrm{Recall}}
\end{align}
where TP, FP, and FN denote the total true positives, false positives, and false negatives, respectively, for the change class.  

\subsection{Framework main results}\label{subsec:framework_results}

\begin{figure}
  \centering
\includegraphics[width=0.99\linewidth]{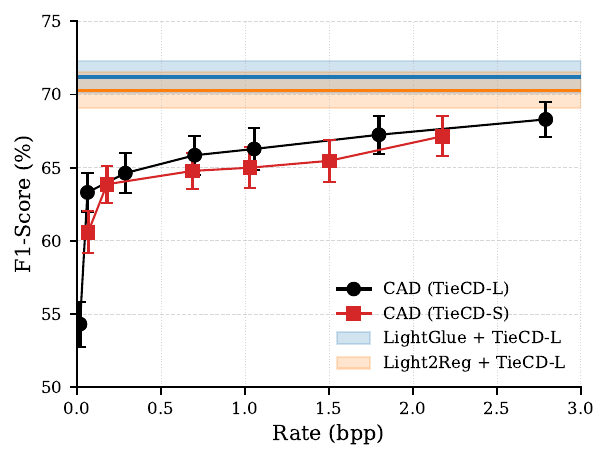}
  \caption{Trade‐off between compression rate and F1-Score. Each marker on the solid curves shows the mean F1‐score (with error bars indicating standard deviation) achieved by the entire \ourmodel\ framework at a given compression rate in bits-per-pixel, employing both \ourcd-S and \ourcd-L. The two horizontal lines correspond to the mean F1‐score of the framework without compression (i.e., \textit{Reg-CD}), with shaded regions showing their respective standard deviations. }
  \label{fig:f1-rate}
\end{figure}

Fig. \ref{fig:f1-rate} shows the main quantitative results for the proposed \ourmodel\ framework. We report the F1 score achieved on the CD task as a function of the rate of the compressed image representations in bits-per-pixel. We evaluate two variants of \ourmodel\ framework, one using the smaller \ourcd-S as CD module and one with the larger \ourcd-L, reporting mean score values and standard deviation error bars accounting for the variability in spatial transformations. The horizontal lines represent benchmark results obtained from experiments without compression, i.e., in\textit{ Reg-CD} configuration, representing a theoretical upper bound to the \ourmodel\ framework in the case where there is no loss due to compression. Two baselines are shown: one uses our proposed \ourreg\ registration network (in orange), which is highly efficient and tailored for onboard deployment; the other uses LightGlue \cite{lindenberger2023lightglue} (in blue), a lighter and more accurate version of SuperGlue \cite{sarlin2020superglue} that effectively represents the state-of-the-art in image matching and registration. We notice that the significantly more expensive LightGlue only allows a modest improvement in F1-score (+0.9, as shown in Table \ref{tab:reg_results}), demonstrating that \ourreg, although much lighter, provides good image co-registration for the CD task. In Subsec. \ref{subsec:registration} we further analyze the differences between LightGlue and \ourreg, also from a computational point of view. We can notice that at a rate of approximately 3 bpp, \ourmodel\ is close to the performance of the uncompressed system. Moreover, a slow, smooth decrease in F1-score is reported as rate is lowered until at very low rates (below 0.1 bpp), F1-Score falls by 10 percentage points. These results suggest that CD can be performed onboard with small requirements in terms of storage as good accuracy can be obtained even for very high compression ratios.

Table \ref{tab:complexity_framework} summarizes the overall computational profile of \ourmodel\ framework. With 121 GFLOPs and 8.8 million parameters, \ourmodel\ (using \ourcd-L) achieves a throughput of 685 Kpixels/s when evaluated on the low-power Nvidia Jetson Orin Nano. With a small loss in terms of CD accuracy, as shown in Fig. \ref{fig:f1-rate}, one can further speed (and lighten) up the pipeline by using the smaller CD network version, \ourcd-S. This smaller variant increases throughput by about 80K pixels/s, and decreases memory usage by about 300MB. Both Fig. \ref{fig:f1-rate} and Table \ref{tab:complexity_framework} demonstrate that the framework comfortably meets the efficiency constraints of onboard deployment.

\begin{table}[t]
    \centering
    \caption{\ourmodel\ computational complexity and throughput. Inference time and memory usage are computed on a Jetson Orin Nano 8GB passing \(3 \times 512 \times 512\) input tensors.}
    \label{tab:complexity_framework}
    \setlength{\tabcolsep}{3pt}
    \begin{tabular}{l c c c c c}
    \toprule
    \textbf{TieCD} & \textbf{Params}  & \textbf{FLOPs}    & \textbf{Inference} & \textbf{Memory} & \textbf{Throughput} \\
             & &  & \textbf{(seconds)}      & \textbf{(MB)} &   \textbf{(pixels/sec.)}     \\
    \midrule
    Large & 8.81M     & 121.47G   & $0.383 \pm 0.001$ & $5729 \pm 9$ &  685K    \\
    Small & 8.22M     & 110.9G   & $0.341 \pm 0.001$ & $5391 \pm 2.37$ &  768K \\ 
    \bottomrule
    \end{tabular}
\end{table}

\subsection{Change detection module results}\label{subsec:cd}

\begin{table}[t]
    \centering
    \caption{F1-Score on Levir‐CD for both input orders.}
    \label{tab:cd_results}
    \setlength{\tabcolsep}{2pt}
    \begin{tabular}{l c c c c}
    \toprule
    \textbf{CD Method} & \textbf{Params} & \textbf{FLOPs} & \textbf{F1}$(x_{t_1},x_{t_2})$ & \textbf{F1}$(x_{t_2},x_{t_1})$ \\
    \midrule
    \textbf{TieCD-S (ours)} & 0.20M   & 15.25G  & 91.04 & 91.04 \\
    \textbf{TieCD-L (ours)} & 0.81M   & 58.98G  & 91.61 & 91.61 \\
    \hline
    TinyCD \cite{codegoni2023tinycd}                &  0.29M  & 11.20G  & 90.33 & 56.74 \\
    FC‐Siam‐Conc \cite{daudt2018fully}              & 1.55M   & 39.93G  & 87.14 &  1.42 \\
    ChangerEx (R18) \cite{fang2023changer}          & 11.4M   & 47.46G  & 91.81 & 91.83 \\
    LightCDNet‐L \cite{xing2023lightcdnet}          &  2.82M  & 50.85G  & 91.15 & 58.00 \\
    ChangeFormer (B1) \cite{bandara2022transformer}& 13.9M   & 52.84G  & 91.22 &  5.08 \\
    HANet \cite{han2023hanet}                       &  3.03M  & 135.8G  & 90.53 &  8.30 \\
    ChangeStar \cite{zheng2021change}               & 17.0M   & 153.5G  & 91.26 & 91.03 \\
    BAN (ViT‐L14‐b1) \cite{li2024new}               & 16.0M   & 574G    & 92.10 &  4.70 \\   
    TTP \cite{chen2023time}                        &  6.21M  & 1.72T   & 91.93 & 86.38 \\
    \bottomrule
    \end{tabular}
\end{table}

We now assess the individual performance of our proposed \ourcd\ module in order to understand its positioning compared to state-of-the-art techniques. In this experiment, images are perfectly registered and no compression is applied. We remark that our design goal is to strike a balance between complexity and detection accuracy. Table \ref{tab:cd_results} reports CD performance of \ourcd\ (in both its small and large variants) and other state-of-the-art CD methods in terms of F1-score. Notice that we report results for both the standard temporal ordering and a swapped ordering of the input image pair. \ourcd\ is the only method whose predictions remain exactly the same when images are swapped, thanks to the strict temporal invariance enforced in the design. ChangeStar \cite{zheng2021change} and ChangerEx \cite{fang2023changer} promote ``temporal symmetry'' by randomly swapping input pairs and computing losses on both orders, and are reasonably robust to ordering. 
TTP \cite{chen2023time} is based on the SAM \cite{kirillov2023segment} foundation model and loses nearly 6 points of F1-score when ordering is swapped, signaling a clear overfitting to ordering. All other methods, however, do not provide any invariance mechanism to temporal order. It can be seen that their performance is totally degraded simply by switching the order of the input images; emblematic is the example of BAN \cite{li2024new}, again a CD method leveraging a foundation model, which achieves the highest F1-Score, but drops to only 4.7\% when the order is changed, effectively making any prediction useless.  Aside from the two large foundation models, \ourcd-L is outperformed only by ChangerEx, reaching state-of-the-art performance. Meanwhile, our efficient \ourcd-S version is less complex than all other methods, except for TinyCD \cite{codegoni2023tinycd} and outperforms it by 0.7 percentage points.

\subsection{Registration module results}\label{subsec:registration}

\begin{figure*}[t]
  \centering
  \includegraphics[width=0.85\linewidth]{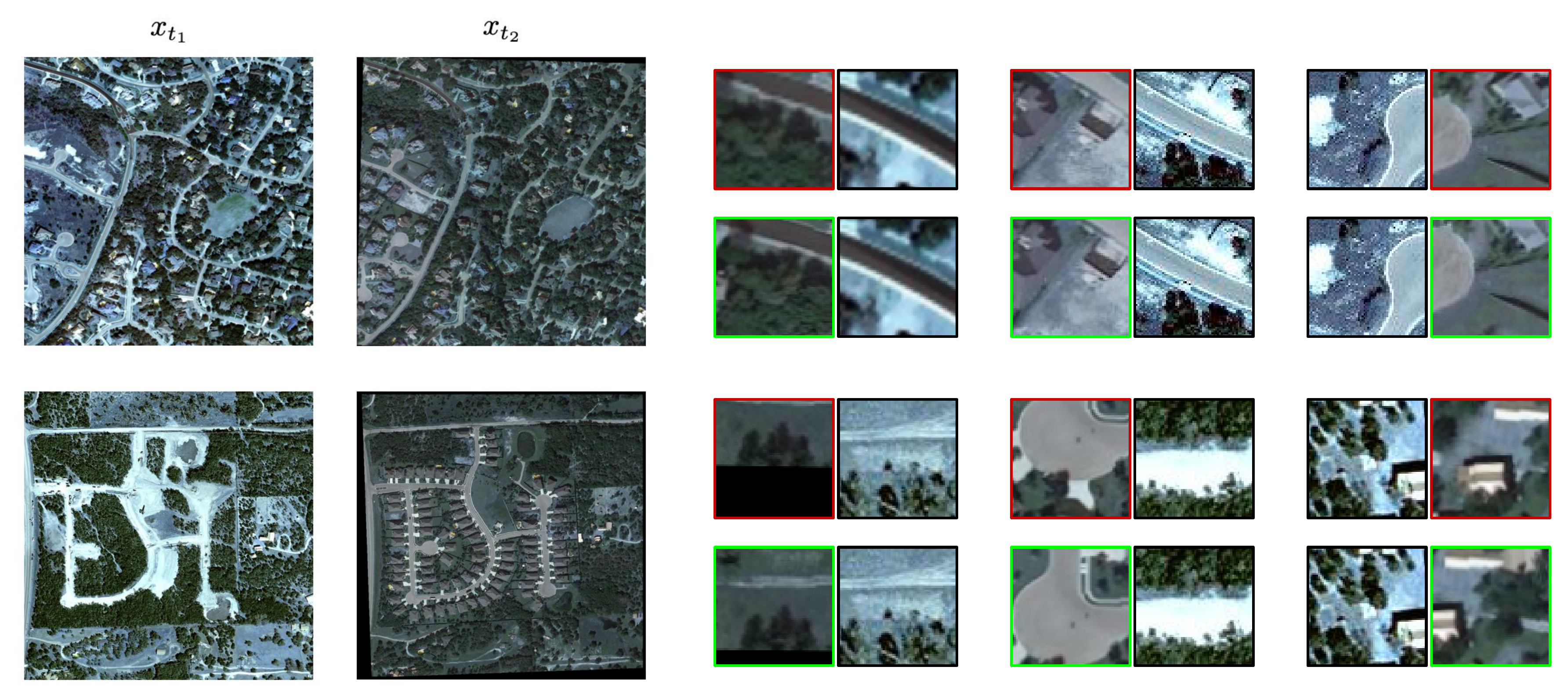}
  \caption{Qualitative co-registration results on two misaligned LevirCD image pairs. For each pair, the \textbf{black border} patches are cropped from the original reference frame \(x_{t_1}\). The \textcolor{red}{\textbf{red border}} patches show the result of warping transformed image \(x_{t_2}\) using our registration network after independent pretraining (\textit{Solo} setup), while the \textcolor{green}{\textbf{green border}} patches show the warp obtained after joint train with the CD module (\textit{Reg–CD} setup). Joint training yields noticeably better alignment, especially around building and roads edges, demonstrating how the downstream CD objective refines the geometric estimates.}
  \label{fig:solo_vs_reg-cd}
\end{figure*}

In this section, we analyze the effectiveness of the registration module in terms of its impact on the performance of the CD task, the improvement due to the joint training strategy and computational complexity against alternative state-of-the-art approaches.

Table \ref{tab:reg_results} reports the CD F1-score without any compression when the registration and CD module are used under two evaluation settings. In the \textit{Solo} setting, each model is simply pretrained (as detailed in Sec. \ref{subsec:training_procedure}) on its own objective and then frozen at test time, allowing us to measure the raw alignment quality of the registration method without further adaptation. In the \textit{Reg–CD} setting, the registration and CD modules are jointly finetuned, providing insight into how much each task benefits from shared optimization.

We also compare our \ourreg\ co-registration network with the state-of-the-art for DL-based keypoint matching: LightGlue \cite{lindenberger2023lightglue}. Unlike \ourreg, which directly regresses a single homography matrix, LightGlue requires first extracting keypoint descriptors and then iteratively matching them. Based on the most accurate results of LightGlue paper, we opt for SuperPoint \cite{detone2018superpoint} as a DL-based method to extract keypoints. 
Notice that Table \ref{tab:reg_results} reports parameter count including both SuperPoint and LightGlue models, while the number of FLOPs is variable due to scene complexity, since iterative algorithms are used, and thus we report a lower bound. 
This lower bound accounts for the two inference passes by SuperPoint (178.5 GFLOPs each), excluding the matching iterations, which we observed ranging from an additional 20 GFLOPs for trivial image pairs to more than 120 GFLOPs for challenging cases.

The experimental results show that unregistered images degrade the CD F1 score, even when state-of-the-art techniques like LightGlue are used. However, we notice how the proposed \ourreg\ in the joint \textit{Reg-CD} training setup has very close performance, despite requiring about one order of magnitude fewer FLOPs. We also notice how the joint \textit{Reg-CD} training setup significantly improves over \textit{Solo} training, which is only able to provide coarse alignment that limits downstream CD accuracy. This is to be expected since the \textit{Solo} training procedure relies on regressing the homography applied to a single image, thus having no content variation in the image pair. However, when used for CD, there are significant content differences. Hence, supplementing the training objective with the CD ground truth provides better alignment, as it trains the model to be more robust to scene variations. Lastly, we observe that also the pipeline using LightGlue benefits from the joint training with the CD model. A qualitative example of the impact of joint \textit{Reg-CD} training  is shown in Fig. \ref{fig:solo_vs_reg-cd}.

\begin{table}[t]
    \centering
    \caption{F1-Score on unregistered LevirCD dataset. Two configurations are proposed: \textit{Reg-CD}, where registration and CD models are jointly finetuned after their independent pretrainings, and \textit{Solo}, where registration and CD models are independently pretrained and tested without further finetuning.}
    \label{tab:reg_results}
    \setlength{\tabcolsep}{3pt}
    \begin{tabular}{l c c c}
    \toprule
    \textbf{Registration method}                     & \textbf{Params}  & \textbf{FLOPs}       & \textbf{F1-Score}         \\
    \midrule
    \ourreg\ (Solo)                             & 1.06M  & 45.18G     & 55.87 (± 1.59)   \\
    \ourreg\ (Reg–CD)                           & 1.06M  & 45.18G     & 70.30 (± 1.23)   \\
    Superpoint + LightGlue (Solo)           & 13.15M & \(>\)357G  & 70.90 (± 0.91)   \\
    Superpoint + LightGlue (Reg–CD)         & 13.15M & \(>\)357G  & 71.20 (± 1.07)   \\
    \bottomrule
    \end{tabular}
\end{table}

\subsection{Compression module results}

\begin{figure}
  \centering
  \includegraphics[width=0.99\linewidth]{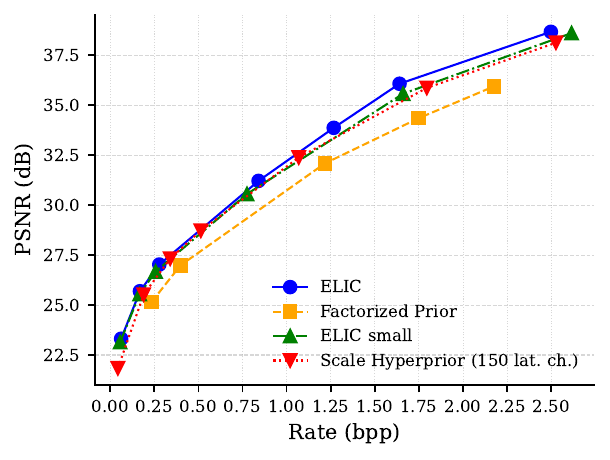}
  \caption{Rate-distortion curves of compression models trained and tested on LevirCD dataset. }
  \label{fig:f1-rate-compression}
\end{figure}

In this section, we perform a comparative evaluation of DL image compression models in order to select one that strikes a balance between rate-distortion performance and low complexity.
Fig. \ref{fig:f1-rate-compression} compares the rate–distortion curves of four DL image compression models trained and evaluated on the LevirCD train split. We compared two well-know methods \cite{balle2018variational}, namely the Factorized Prior model and a custom reduced variant of the Scale Hyperprior mode with only 150 latent channels, as well as the newer state-of-the-art method ELIC \cite{he2022elic} and its smaller variant. The Factorized Prior consistently lags behind the others at both low and high bitrates, yielding unsatisfactory reconstruction quality and was therefore excluded from further consideration. ELIC achieves the highest PSNR among the four models, but at the expense of a large model size and FLOPs count (see Table \ref{tab:comp_complexity}). The smaller version of ELIC reduces computational complexity, yet still requires more than twice the FLOPs of the reduced variant of the Scale Hyperprior. Finally, the reduced variant of the Scale Hyperprior delivers a competitive rate-distortion curve, closely matching ELIC small and approaching standard ELIC, while using only 62G FLOPs and 7M parameters, making it the best choice for our onboard compression module. 

\begin{table}[t]
    \centering
    \caption{Complexity of compression models.}
    \label{tab:comp_complexity}
    \setlength{\tabcolsep}{3pt}
    \begin{tabular}{l c c}
    \toprule
    \textbf{Compression method}                                       & \textbf{Params}   & \textbf{FLOPs}     \\
    \midrule
    Custom Scale Hyperprior \cite{balle2018variational}             & 6.96M   &  62.23G  \\
    Factorized Prior \cite{balle2018variational}             & 3.00M   &  44.28G  \\
    ELIC (small) \cite{he2022elic}                           & 26.63M  & 168.82G  \\
    ELIC \cite{he2022elic}                                   & 35.42M  & 314.56G  \\
    \bottomrule
    \end{tabular}
\end{table}

\subsection{\ourmodel\ complexity and performance ablations}

\begin{figure}
  \centering
  \includegraphics[width=0.99\linewidth]{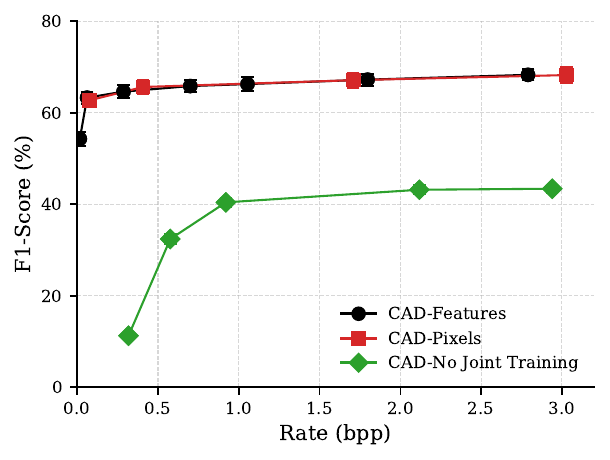}
  \caption{Rate-F1 curves for \ourmodel\ framework (\ourmodel-Features), \ourmodel\ framework variant working in pixel space (\ourmodel-Pixels) and a baseline where all the modules are just independently trained, each one for its specific task, without finetuning (\ourmodel-No Joint Training). }
  \label{fig:f1-rate-comparison}
\end{figure}

\begin{table}[t]
    \centering
    \caption{\ourmodel\ computational complexity in pixel and feature spaces. Inference time and memory usage are computed on a Jetson Orin Nano 8GB. All the metrics refer to inference using the entire \ourmodel\ framework, including all three submodules.}
    \label{tab:complexity_framework_comparison}
    \setlength{\tabcolsep}{3pt}
    \begin{tabular}{l c c c c}
    \toprule
    \textbf{Domain}   & \textbf{Params}  & \textbf{FLOPs}    & \textbf{Inference Time} & \textbf{Memory Usage} \\
             &  &  & \textbf{(seconds)}      & \textbf{(MB)}         \\
    \midrule
    Pixel    & 8.83M     & 166.39G   & $0.734 \pm 0.001$  & $6829 \pm 8$      \\
    Feature  & 8.81M     & 121.47G   & $0.383 \pm 0.001$  & $5729 \pm 9$      \\
    \bottomrule
    \end{tabular}
\end{table}

Fig. \ref{fig:f1-rate-comparison} and Table \ref{tab:complexity_framework_comparison} analyze the trade-offs between three variants of \ourmodel\ framework. The black curve (\ourmodel-Features) corresponds to the standard configuration (with \ourcd-L) already presented in the main results in Fig. \ref{fig:f1-rate}. 

As described in Sec. \ref{subsec:framework}, \ourmodel\ operates primarily in the feature space, with the sole exception of the registration module, which receives the full-resolution images to regress the homography. The \ourmodel-Pixels (red curve) design presents a variant in which the CD module uses an input in the full-resolution pixel space. This is more expensive as it requires extra layers and operations, which are not needed when directly passing deep features from the compression module, resulting in a considerable increase in computational complexity of over 45 GFLOPs and nearly twice the inference time. However, as seen in Fig. \ref{fig:f1-rate-comparison}, the gain in F1-score is negligible, with both curves nearly overlapping across all rates.

Additionally, we analyze the performance when no end-to-end finetuning is performed, i.e., we rely on independent modules pretrained for the compression, registration and change detection tasks. This is reported in Fig. \ref{fig:f1-rate-comparison} as the \ourmodel-No Joint Training variant (green curve). It can be noticed that this approach leads to significantly degraded performance, with low F1-scores even at high rates. This highlights the importance of joint training: simply combining independently optimized modules is insufficient to achieve effective onboard CD.

\subsection{Effectiveness of Pretraining}

\begin{figure}[t]
  \centering
  \includegraphics[width=0.99\linewidth]{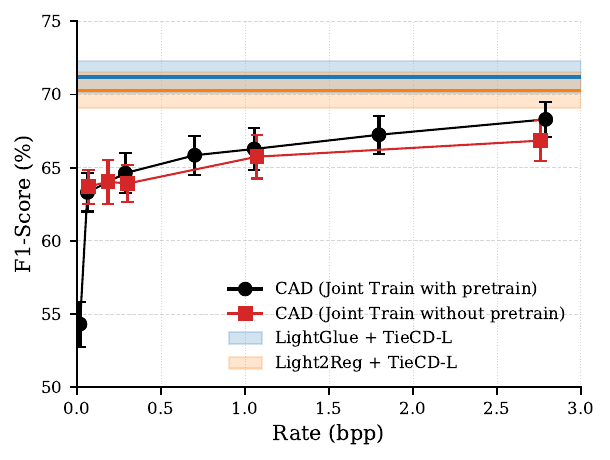}
  \caption{Rate–F1 curves comparing \ourmodel\ with initial module-wise pretraining (black) against the same framework trained end-to-end from scratch (red).}
  \label{fig:pretrain_vs_scratch}
\end{figure}

To evaluate the impact of our proposed pretraining strategy (Sec. \ref{sec:method}), we compare two training regimes: (1) pretraining each module individually before joint fine-tuning the whole \ourmodel\ framework, and (2) training the entire \ourmodel\ framework from random initialization using only the combined loss of Eq.~\eqref{eq:full_loss}. As shown in Figure \ref{fig:pretrain_vs_scratch} the pretrained variant consistently outperforms the scratch baseline at all bitrates, achieving higher F1-scores for a given bpp. In addition, the curve from the pretraining basline is smoother, indicating stable behavior. In contrast, without any pretraining strategy \ourmodel\ shows larger fluctuations and slower, less reliable convergence, underlining the practical importance of our modular pretraining approach.

\section{Conclusions}

We presented a novel framework that for the first time addresses the problem of change detection directly onboard satellites. We showed that issues like the lack of image registration and the limited storage available onboard require the development of an integrated framework that jointly considers compression, registration and CD. By developing a single lightweight neural network model for the entire framework and optimizing it end-to-end, we showed that compelling CD performance at significant compression ratios can be achieved. Importantly, we have found that the joint training allows to employ a low-complexity co-registration module, allowing to obtain a low-complexity design that is suitable for fast inference on low-power devices and has been experimentally tested on one such device.

We believe that with this work we have taken a first big step in the direction of onboard real-time processing. We strongly encourage the scientific community to release raw non-orthorectified image datasets, to further evolve this area of research.

\small
\bibliographystyle{IEEEtran}
\bibliography{biblio}

\begin{IEEEbiography}[{\includegraphics[width=1in,height=1.25in,clip,keepaspectratio]{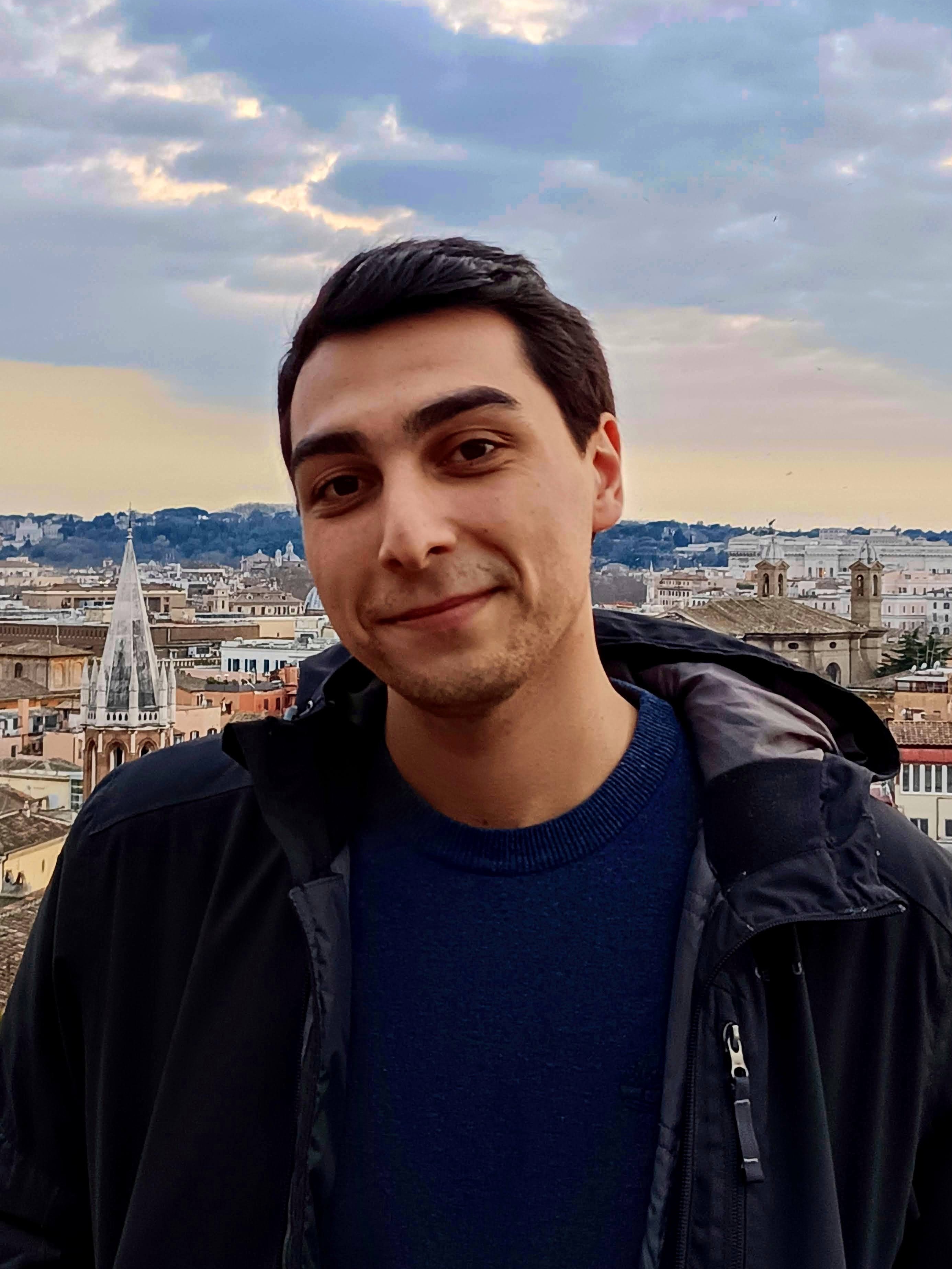}}]{Gabriele Inzerillo}  received the M.Sc. in Computer Engineering from the Politecnico di Torino in 2022. He is currently pursuing a Ph.D. on deep learning for remote sensing, as a result of a collaboration between Politecnico di Torino and the European Space Agency, within the OSIP Framework. \end{IEEEbiography}

\begin{IEEEbiography}[{\includegraphics[width=1in,height=1.25in,clip,keepaspectratio]{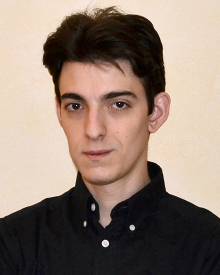}}]{Diego Valsesia} (S'13-M'17) received the Ph.D. degree in electronic and communication engineering from the Politecnico di Torino, in 2016. He is currently an Assistant Professor with the Department of Electronics and Telecommunications (DET), Politecnico di Torino. His main research interests include processing of remote sensing images, and deep learning for inverse problems in imaging. He is a Senior Area Editor for the IEEE Transactions on Image Processing, for which he received the 2023 Outstanding Editorial Board Member Award. He is a member of the EURASIP Technical Area Committee for Signal and Data Analytics for Machine Learning and a member of the ELLIS society. He was the recipient of the IEEE ICIP 2019 Best Paper Award, the IEEE Multimedia 2019 Best Paper Award. \end{IEEEbiography}

\begin{IEEEbiography}[{\includegraphics[width=1in,height=1.25in,clip,keepaspectratio]{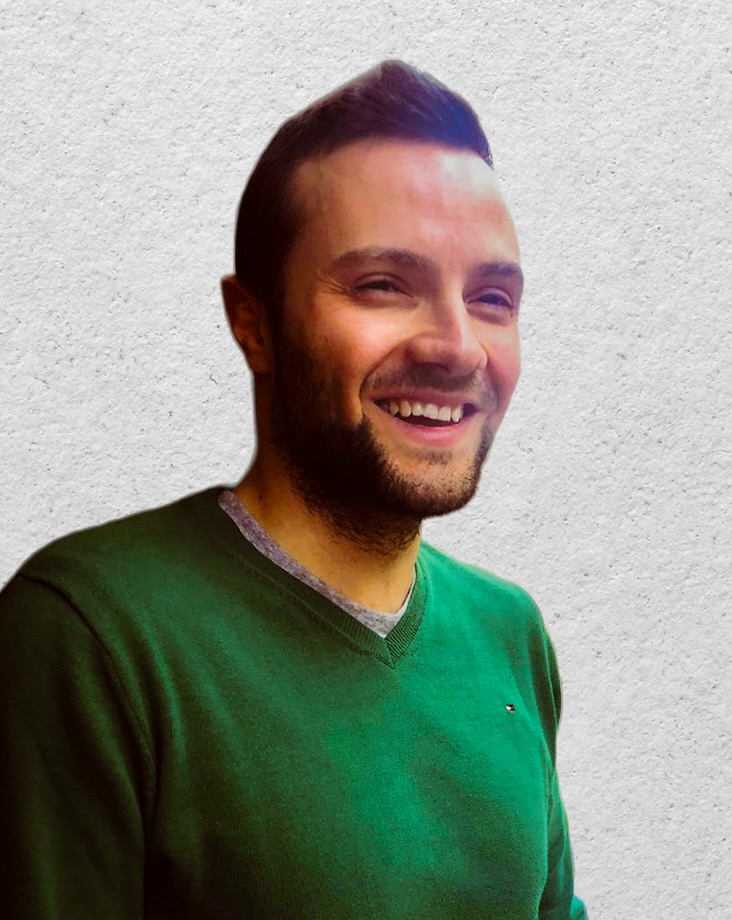}}]{Aniello Fiengo} received a degree in Telecommunication Engineering from the University of Naples Federico II and a PhD in Image and Signal Processing from Institut Mines-Télécom Télécom ParisTech, Paris, in 2016. Currently, at ESA-ESTEC, they oversee the definition and implementation of onboard processing algorithms for Earth Observation missions and serve as Technical Officer for multiple R\&D activities. \end{IEEEbiography}

\begin{IEEEbiography}[{\includegraphics[width=1in,height=1.25in,clip,keepaspectratio]{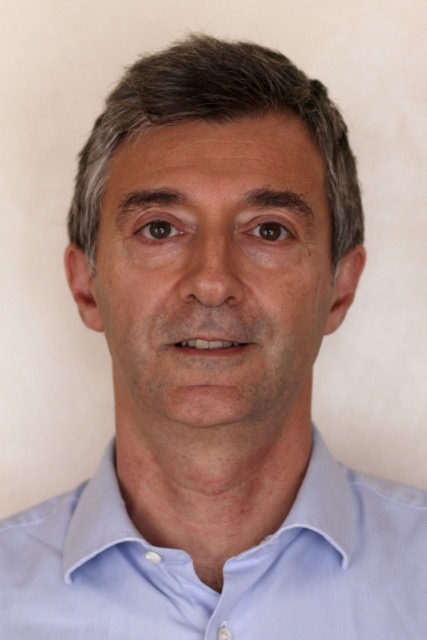}}]{Enrico Magli} (S'97-M'01-SM'07-F'17)(IEEE Fellow) received the M.Sc. and Ph.D. degrees from the Politecnico di Torino, Italy, in 1997 and 2001, respectively. He is currently a Full Professor with Politecnico di Torino, Italy, where he leads the Image Processing and Learning group, performing research in the fields of deep learning for image and video processing, image compression and image forensic for multimedia and remote sensing applications. He is a Senior Associate Editor of IEEE Journal on Selected Topics in Signal Processing, and a former Associate Editor of the EURASIP Journal on Image and Video Processing, the IEEE TRANSACTIONS ON MULTIMEDIA and the IEEE TRANSACTIONS ON CIRCUITS AND SYSTEMS FOR VIDEO TECHNOLOGY. He is a Fellow of the IEEE, a Fellow of the ELLIS Society for the advancement of artificial intelligence in Europe, and has been an IEEE Distinguished Lecturer from 2015 to 2016. He was the recipient of the IEEE Geoscience and Remote Sensing Society 2011 Transactions Prize Paper Award, the IEEE ICIP 2015 Best Student Paper Award (as senior author), the IEEE ICIP 2019 Best Paper Award, the IEEE Multimedia 2019 Best Paper Award, and the 2010 and 2014 Best Associate Editor Award of the IEEE TRANSACTIONS ON CIRCUITS AND SYSTEMS FOR VIDEO TECHNOLOGY.
\end{IEEEbiography}

\end{document}